\documentclass{article}
\usepackage{ijcai17}

\usepackage{times}

\usepackage{physics}
\usepackage{amsmath}
\usepackage{amssymb}
\usepackage{amsthm}
\usepackage{array}
\usepackage{enumitem}
\usepackage{graphicx}
\usepackage{multirow}
\usepackage{subfigure}
\usepackage{placeins}

\usepackage{color}

\usepackage{pifont}

\usepackage[small]{caption}
\newtheorem{definition}{Definition}

\graphicspath{ {figure/} }

\title{Training Group Orthogonal Neural Networks with Privileged Information}

\author{Yunpeng Chen$^{1}$, Xiaojie Jin$^{1}$, Jiashi Feng$^{1}$, Shuicheng Yan$^{2,1}$  \\
$^{1}$National University of Singapore \\
$^{2}$Qihoo 360 AI Institute\\
\{chenyunpeng, xiaojie.jin\}@u.nus.edu, elefjia@nus.edu.sg, \\
yanshuicheng@360.cn \\
}

\begin{document}

\maketitle

\begin{abstract}
Learning rich and diverse representations is critical for the performance of deep convolutional neural networks (CNNs). In this paper, we consider how to use privileged information to promote inherent diversity of a single CNN model such that the model can learn better representations and offer stronger generalization ability. To this end, we propose a novel group orthogonal convolutional neural network (GoCNN) that learns untangled  representations within each layer by exploiting provided privileged information  and enhances representation diversity effectively. We take image classification as an example where image segmentation annotations are used as privileged information during the training process. Experiments on two  benchmark datasets~\textendash~ImageNet and PASCAL VOC~\textendash~clearly demonstrate the strong generalization ability of our proposed GoCNN model. On the ImageNet dataset, GoCNN improves the performance of state-of-the-art ResNet-152 model by absolute value of 1.2\% while only uses  privileged information of 10\% of the training images, confirming effectiveness of GoCNN on utilizing available privileged knowledge to train better CNNs.
\end{abstract}

\section{Introduction}

In many recent works~\cite{he2015deep,szegedy2016inception,zhang2016polynet}, it has been observed that CNNs with different architectures or even different weight initializations may learn slightly different feature representations. Combining these heterogeneous models can provide richer and more diverse feature representation which can further boost the final performance. However, principled methods to directly enhance model inherent diversity are still rare. In this work, we propose a novel method to enhance such feature diversity by exploiting \emph{privileged information} during the training process.

The privileged information~\cite{vapnik2009new} refers to the information that is available at training but not at testing. For the image classification task, the privileged information can be with many different types, such as text description~\cite{vapnik2015learning} and feature vector~\cite{lapin2014learning}. Using privileged information to assist the learning process has been well studied in the area of training a better classifer~\cite{vapnik2009new,lapin2014learning,feyereisl2014object,vapnik2015learning} but still rare in deep learning. In this work, we focus on learning more diverse feature representations. We aim to employ the privileged information during training to obtain a \emph{single} CNN model with sufficient inherent diversity, such that the model learns more diverse representations and has a stronger generalization ability than vanilla CNNs. To this end, we propose a group orthogonal convolutional neural network (GoCNN) model based on the idea of learning different groups of convolutional functions which are ``orthogonal'' to the ones in other groups, \emph{i.e.}, with no significant correlation among the produced features. Optimizing orthogonality among convolutional functions reduces the redundancy and increases the diversity within the architecture.

Properly defining the groups of convolutional functions in the GoCNN is not an easy task. In this work, we propose to exploit available \emph{privileged information} for identifying the proper groups. Specifically, in the context of image classification, object segmentation annotations which are (partially) available in several public datasets give richer information. In addition, the background contents are usually independent on foreground objects within an image. Thus, splitting convolutional functions into different groups and enforcing them to learn features from the foreground and background separately can help construct orthogonal groups with small correlations. Motivated by this, we introduce the GoCNN architecture which explores to learn discriminative features from foreground and background separately where the foreground-background segregation is offered by the privileged segmentation annotation for training GoCNN. In this way, inherent diversity of the GoCNN can be explicitly enhanced. Moreover, benefiting from pursuing the group orthogonality, the learned convolutional functions within GoCNN are demonstrated to be foreground and background diagnostic even for extracting features from new images in the testing phase.

To the best of our knowledge,  this work is the first to explore a principled way to train a deep neural network with desired inherent diversity and the first to investigate how to use the segmentation privileged information to assist image classification within a deep learning architecture. Experiments on ImageNet and PASCAL VOC clearly demonstrate GoCNN improves upon vanilla CNN models significantly, in terms of classification accuracy. Moreover, as a by-product of implementing GoCNN, we also provide positive answers to the following two prominent questions about image classification: 
(1) Does background information indeed help object recognition in deep learning? 
(2) Can a more precise annotation with richer information, \emph{e.g.}, segmentation annotation, assist the image classification training process non-trivially?

\section{Related Work}

Learning rich and diverse feature representations is always desired while training CNNs for gaining stronger generalization ability. Most existing works tend to implicitly pursue such diversity by increasing the CNN's learning capacity~\cite{simonyan2014very,he2015deep,jin2015deep} or design better cost functions~\cite{tang2013deep}. Methods that explicitly encourage inherent feature diversity are still rare so far.
Knowledge distillation~\cite{hinton2015distilling} proposed to process compresses knowledge and thus encourages a weak model to learn more diverse and discriminative features. However, it works in two stages and relys pre-trained complicated teacher network model, which introduce undesired computation overhead. Similar works, \emph{e.g.} the {Diversity Networks}~\cite{srageometric}, aims squeeze the knowledge by preserving the most diverse features to avoid harming the performance. More recently, \cite{cogswell2015reducing} proposed the DeCov approach to explicitly encourage feature diversity which is consistent with our motivation. However, DeCov penalizes the covariance in an unsupervised fashion and cannot utilize extra available annotations, leading to insignificant performance improvement over vanilla models \cite{cogswell2015reducing}. 

Using privileged information to learn better features during the training process is similar in spirit with our method. Both our proposed method and~\cite{lapin2014learning} introduce privileged information to assist the training process. However, almost all existing  works~\cite{lapin2014learning,lopez2016unifying,sharmanska2014learning} are based on SVM$^+$ which only focuses on training better classifier and is not able to do the end-to-end training for better features.

It is also worth to notice that simply adding a segmentation loss to image classification neural network is not equivalent to a GoCNN model. This is because image segmentation requires each pixel within the target area to be activated and the others stay silent for dense prediction, while GoCNN does not require the each pixel within the target area to be activated. GoCNN is specifically designed for classification tasks, not for segmentation ones. Moreover, our proposed GoCNN supports learning from partial privileged information while the CNN above needs a fully annotated training set.

\section{Model Diversity of Convolutional Neural Networks} 
Throughout the paper, we use $c^{(k)}$ to denote the total number of convolutional functions (or filters) at the $k$-th layer and use $i$ and $j$ to index different functions, \emph{i.e.}, $f_i^{(k)}(\cdot)$ denotes the $i$-th convolutional function at the $k$-th layer of the network. The function $ f $ maps an input feature map to another new feature map. The height and  the width of a feature map output by the layer $ k $ are denoted as $h^{(k)}$ and $w^{(k)}$ respectively. We consider a network model consisting of $N$ layers in total. 

Under a standard CNN architecture,  elements within the $i$-th feature map output by the layer $k$  are produced by the 
same convolutional function $f_i^{(k)}$. Thus,  these elements  represent response of the input w.r.t.\ the same type of features across different locations. Therefore, encouraging the feature to be diverse within a single 
feature map does not make sense. In this work, our target is to enhance the diversity among different convolutional functions.
Before presenting details,  we  give a formal description  of \textit{model diversity} for an  $N$-layer CNN.

\begin{definition}[Model Diversity]
	Let $f_i^{(k)}$ denote the $i$-th convolutional function at the $k$-th layer of a neural network model, and then the model diversity of the $k$-th layer is defined as
\begin{equation}
\label{eqn:diverse_def}
\zeta^{(k)} \triangleq 1 - \frac{1}{{c^{(k)}}^2  } 
\sum_{i,j=1}^{c^{(k)}} \abs{\operatorname{corr}\left(f^{(k)}_{i}, f^{(k)}_{j}\right)}.
\end{equation}
Here the operator $\operatorname{corr}(\cdot,\cdot)$ denotes the statistical \emph{correlation} w.r.t.\ the training samples. 
\end{definition}

In other words, the inherent diversity of a network model that we are going to maximize is evaluated across all the convolutional functions within the same layer.

Directly 
maximizing the value of $\zeta^{(k)}$ during training the network is quite involved
 due to the large combination number of different convolutional functions. 
To address the difficulty, we propose to 
maximize diversity by learning the convolutional functions in different \emph{groups} independently. The functions from different groups are untangled  from each other naturally and their correlation would be small. Suppose the convolutional functions at each layer are partitioned into $ m $ different groups, denoted as $ \mathcal{G} = \{G_1,\ldots,G_m\} $. Then, we propose to maximize the following \textit{Group-wise Model Diversity}.
\begin{definition}[Group-wise Model Diversity]
	Given a pre-defined group partition set $ \mathcal{G} = \{G_1,\ldots,G_m\} $ of convolutional functions at  a specific layer, the group-wise model diversity of this layer is defined as
	\begin{equation*}
	\zeta_g^{(k)} \triangleq 1 - \frac{1}{Z} 
	\sum_{s\not=t}^{|\mathcal{G}|}\sum_{i\in G_s, j \in G_t}\abs{ \operatorname{corr}\left(f_i^{(k)} , f_j^{(k)}\right)}.
	\end{equation*}
{Here $Z$ is the normalization factor that ensures the  normalized summation to be not greater than one.}
\end{definition}

Instead of directly optimizing the \textit{model diversity}, we consider optimizing the \textit{group-wise model diversity} by finding a set of orthogonal groups $\{G^*_1,\ldots,G^*_m\}$, where convolutional functions within each group are uncorrelated with others within different groups. In the scenario of image representation learning, one typical example of such orthogonal groups is the foreground group and background group pair~\textemdash~partitioning the functions into two groups and letting them learn features from foreground and background contents respectively.

\begin{figure*}[t]	
\center
	\vspace{-1mm}
	\resizebox{1\textwidth}{!}{	
	 \includegraphics{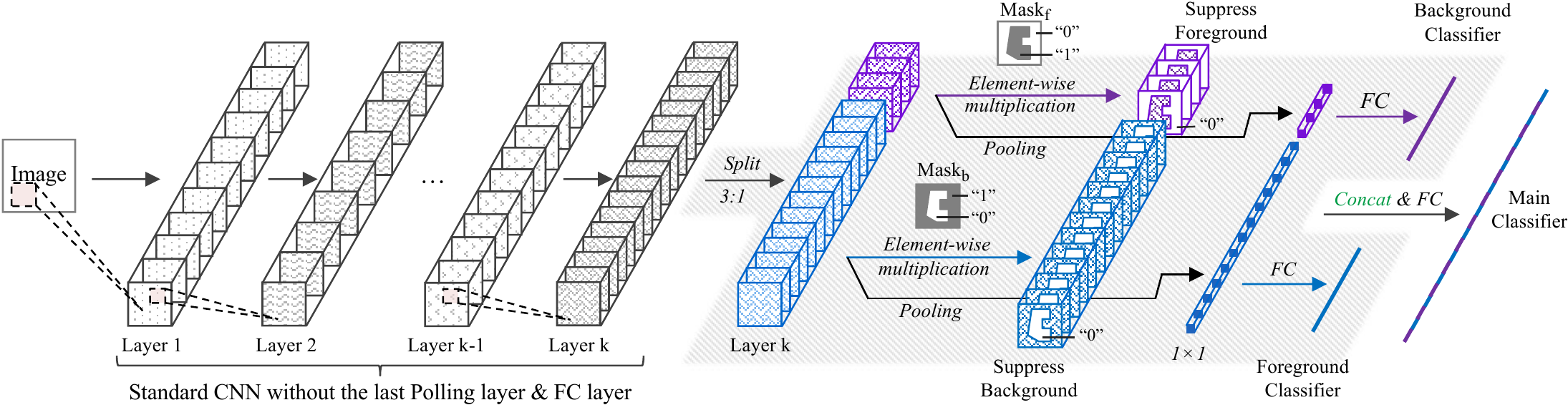}
	}	
	\vspace{-5mm}
	\caption{Architecture of the GoCNN. GoCNN is built upon a standard CNN architecture where the final convolution layer are explicitly divided into two groups: the foreground group (blue) which concentrates on learning the foreground feature and the background group (purple) which learns the background feature. The output features of these two groups are concatenated as a whole representation of the input image. In testing phase, parts within the gray shadow are removed and the ``Concat'' (green) operation is replaced by a ``Pooling'' operation making GoCNN back to a standard CNN.}
	\label{mainframework}
\vspace{-2mm}
	
\end{figure*}
\FloatBarrier

In this work, we use segmentation annotation as privileged information for finding orthogonal groups of convolutional functions $ G^*_1,\ldots, G^*_m $. In particular, we derive the foreground and background segregation from the privileged information for an image. Then we partition convolutional functions at a specific layer of a CNN model into \emph{foreground} and \emph{background} groups respectively, and train a GoCNN model to learn the foreground and background features separately.

\section{Group Orthogonal Convolutional Neural Networks} 

We introduce the group orthogonal constraint to maximize group-wise diversity among different groups of convolutional functions explicitly by constructing a group orthogonal convolutional neural network (GoCNN). Details on the architecture of GoCNN are shown in Figure~\ref{mainframework}.

\vspace{-1mm}
\subsection{Groups Construction}
\vspace{-0mm}

To learn convolutional functions that are  specific for foreground content of an image, we propose the following two constraints for the foreground group of functions. The first constraint forces the functions to be learned from the foreground only and free of any contamination from the background, and the second constraint encourages the learned functions to be discriminative for image classification. 

We learn features that only lie in the foreground by suppressing any contamination (the information we want to avoid) from the background. As aforementioned,  we use the object segmentation annotations (denoted as $ \mathrm{Mask} $ that takes a binary value at each image location) as the privileged information in the training phase. The  $ \mathrm{Mask} $  is used to  identify   background features where the  foreground convolutional functions should not be responsive. 

To implement this, we introduce an extractor that is adopted on each feature map within the foreground group to extract function response on the background. In particular, we define an extractor $\varphi(\cdot,\cdot)$  as follows:
\begin{equation}
	\label{eq-filter}
	\varphi(f^{(k)}_i(x), \mathrm{Mask} ) \triangleq f^{(k)}_i(x) \odot \mathrm{Mask},
\vspace{-0.5mm}
\end{equation}
where $x$ denotes the raw input and $\odot$ denotes the element-wise multiplication.

In the above operator, we use the background object mask $\mathrm{Mask}_b$ to extract background features. Each element in $\mathrm{Mask}_b$ is equal to one if the corresponding position lies on  background and zero otherwise. 
The extracted background features are then  suppressed by the following regression loss (the regression target is 0): 
\begin{equation}
\label{eq-supprese-fg}
\min_{\theta} \sum_{i} \| \varphi(f^{(k)}_i(x;\theta), \mathrm{Mask}_b)\|_F.
\end{equation}
Here $\theta$ parameterizes the convolution function $ f^{(k)}_i $. Since the target value for this regression is zero, we also call it a \emph{suppression} term. It will only suppress the response output by $f^{(k)}_i$ at the locations outside the foreground.

For the second constraint, \emph{i.e.}, encouraging the functions to learn discriminative features, we simply use the standard softmax classification loss to supervise the learning phase.

The role of the background group is complementary to the foreground one. It aims to learn convolutional functions that are only specific for background contents. Thus, the functions within the background group have a same suppression term as in Eqn.~\eqref{eq-supprese-fg}, in which  $\mathrm{Mask}_b$ is replaced with $\mathrm{Mask}_f$ to restrict the learned features to make them only lie in the background space. The $\mathrm{Mask}_f$ is simply computed as $\mathrm{Mask}_f = 1 - \mathrm{Mask}_b$.
Also, a softmax linear classifier is attached during training to guarantee that these learned background functions are useful for predicting image categories.

\subsection{ Architecture and Implementation Details of The GoCNN}

In GoCNN, the size ratio of foreground group and background group is fixed to be $3{:}1$ during training, since intuitively the foreground contents are much more informative than the background contents in classifying images. As shown in Figure~\ref{mainframework}, a single fully connected layer (or multiple layers depending on the basic CNN architecture) is used to unify the functional learning within different groups and combine features learned from different groups. It aggregates the information from different feature spaces and produces the final image category prediction. During the testing stage, parts unrelated to the final main output will be removed. Therefore, in terms of testing, neither extra parameters nor extra computational cost is introduced. The GoCNN is exactly the same as the adopted CNN in the testing phase.

In summary, for an incoming training sample, it passes through all the layers to the final convolution layer. Then the irrelevant features for each group (foreground or background) will be filtered out by privileged segmentation masks. Those filtered features will then flow into a suppressor (see Eqn.~\eqref{eq-supprese-fg}). For the output features from each group, it will flow up along two paths: one leads to the group-wise classifier, and the other one leads to the main classifier. The three gradients from the suppressors, the group-wise classifiers and the main classifier will be used for updating the network parameters.

\vspace{-4mm}

\paragraph{Applications with Incomplete Privileged Information} 
Our proposed GoCNN can also be applied for semi-supervised learning. When only a small subset of images have the privileged segmentation annotations in a dataset, we simply set the segmentations of images without annotations to be $\mathrm{Mask}_f = \mathrm{Mask}_b = \mathbf{0}$, where $\mathbf{0}$ is the matrix with all of its elements being $0$. In other words, we disable both the suppression terms (ref. Eqn.~\eqref{eq-supprese-fg}) on foreground and background parts as well as the extractors on the back propagation path. By doing so, fully annotated training samples with privileged information will supervise GoCNN to learn both discriminative and diverse features while the samples with only image tags only guide GoCNN to learn category discriminative features.

\section{Experiments}

\subsection{Experiment Settings and Implementation Details}

\paragraph{Datasets} 
We evaluate the performance of GoCNN in image classification on two benchmark datasets, \emph{i.e.}, the ImageNet~\cite{deng2009imagenet} dataset and the Pascal VOC 2012	dataset~\cite{everingham2010pascal}.

\begin{itemize}[leftmargin=0.15in]
	\item \textbf{ImageNet } \space \space
	ImageNet dataset contains 1,000 classes with about 1,300 images for each class and 1.2 million images in total, but without any image segmentation annotations. To collect privileged information, we randomly select 130 images from each class and manually annotate the object segmentation masks for them, resulting in 130k training images, which we refer to as ``ImageNet-0.1m''. By merging the ``ImageNet-0.1m'' with the original ImageNet datatset, we get a blended dataset with 1.2 million training images and 10\% privileged information, which we call ``ImageNet-Full''. We use the original validation set of ImageNet for evaluation. Note that neither our baselines nor the proposed GoCNN needs segmentation information in testing phase. 
	
	\item \textbf{PASCAL VOC 2012} \space \space
	The PASCAL VOC 2012 dataset contains 11,530 images from 20 classes. For the classification task, there are 5,717 images for training and 5,823 images for validation. We use this dataset to further evaluate the generalization ability of different models including GoCNN trained on the ``ImageNet-0.1m'': we pre-train the evaluated models on the ``ImageNet-0.1m'' dataset and fine-tune them on PASCAL VOC 2012 training set. We evaluate their performance on the validation set.
	
\end{itemize}

\vspace{-1mm}

\paragraph{The Basic Architecture of GoCNN} 
In our experiments, we use the ResNet~\cite{he2015deep} as the basic architecture to build GoCNN. Since our focus is on justifying the effectiveness of our proposed method, rather than pushing the state-of-the-art, and the deepest ResNet contains 152 layers which will cost weeks to train, we choose a light version of architecture (ResNet-18~\cite{he2015deep}) that contains 18 layers as our basic model for most experiments. The ResNet-152~\cite{he2015deep} is only used for experiment on the ``ImageNet-Full''. The used loss function for the single class classification on ImageNet dataset is the standard softmax loss. When performing multi-label classification on \textit{PASCAL VOC}, we use the logistic regression loss.

\vspace{-1mm}

\paragraph{Training and Testing Strategy} 
We use MXNet~\cite{chen2015mxnet} to conduct model training and testing. We train GoCNN from scratch. Images are resized with a shorter side randomly sampled within [256, 480] for scale augmentation and $224\times 224$ crops are randomly sampled during training~\cite{he2015deep}. We use SGD with base learning rate equal to 0.1 at the beginning and reduce the learning rate by a factor of 10 when the validation accuracy saturates. For the experiments on ResNet-18 we use single node with a mini-batch size of 512. For the ResNet-152 we use 48 GPUs with mini-batch size of 32 per GPU. Following~\cite{he2015deep}, we use a weight decay of 0.0001 and a momentum of 0.9 in the training. 
We evaluate the performance of GoCNN on two different testing settings: the complete privileged information setting and the partial privileged information setting. We perform 10-crop testing~\cite{he2015deep} for the complete privileged information scenario, and do a single crop testing~\cite{he2015deep} for the partial privileged information scenario for convenience.

\vspace{-1mm}
	 
\paragraph{Compared Baseline Models} 
Our proposed GoCNN follows the Learning Using Privileged Information (LUPI) paradigm~\cite{lapin2014learning}, which exploits additional information to facilitate learning but does not require extra information in testing. There are a few baseline models falling into the same paradigm. 
One is the SVM+ method~\cite{pechyony2011fast} and the other one is the standard model (\emph{i.e.}, the ResNet). We simply refer to ResNet-18 by \textit{baseline} if no confusion occurs. In the experiments, we implement the SVM+ using the code provided by~\cite{pechyony2011fast} with default parameter settings and linear kernel. We follow the scheme as described in~\cite{lapin2014learning} to train the SVM+ model. More concretely, we train multiple one-versus-rest SVM+ models upon the deep features extracted from both the entire images and the foreground regions (used as the privileged information). We use the averaged pooling over 10 crops on the feature maps before the $FC$ layer as the deep feature for training SVM+. It is worth noting that all of these models (including SVM+ and GoCNN) use a linear classifier and thus have the same number of parameters, or more concretely, GoCNN does not require more parameters than SVM+ and the vanilla ResNet.

\renewcommand{\arraystretch}{1.1}
\begin{table*}[t]
\setcounter{table}{1}
\centering
	\caption{Validation accuracy (for 10-crop validations) of different components of 
			 GoCNN on ImageNet validation set. \emph{ResNet-18-obj} refers 
			 to the baseline model trained on  \emph{pure object} ``ImageNet-0.1m'' dataset, 
			 \textit{Only Bg} refers to our proposed model with foreground part 
			 gradient blocked, and \textit{Only Fg} refers to our proposed model 
			 with background part gradient blocked. ($^*$ marks the 
			 part which shares the same classifier with the main classifier.) }
	\vspace{-2mm}	
	\label{ImageNet-Complete2}
	\resizebox{1\textwidth}{!}{
	\tiny
	\begin{tabular}{c|c|c|c|c|c|c}
	\hline
			& \multicolumn{3}{c|}{Top-1 Accuracy (\%)}                                                                                                                         
			& \multicolumn{3}{c }{Top-5 Accuracy (\%)}                                                                                                                                                     
	\\ \hline
	        & \begin{tabular}[c]{@{}c@{}}Main\_classifier\end{tabular} 
	        & \begin{tabular}[c]{@{}c@{}}Fg\_classifier\end{tabular} 
	        & \begin{tabular}[c]{@{}c@{}}Bg\_classifier\end{tabular} 
	        & \begin{tabular}[c]{@{}c@{}}Main\_classifier\end{tabular} 
	        & \begin{tabular}[c]{@{}c@{}}Fg\_Classifier\end{tabular} 
	        & \begin{tabular}[c]{@{}c@{}}Bg\_Classifier\end{tabular} 
	\\ \hline
	ResNet-18-obj
			& {12.45} 				& 	12.45$^*$			& ---~
			& {24.43}				&	24.43$^*$			& ---~
	\\ 
	\begin{tabular}[c]{@{}c@{}}Only Bg\end{tabular} 
			& {40.36} 				& 	---~					&  40.36$^*$
			& {67.24}				&	---~					&  67.24$^*$
	\\ 
	\begin{tabular}[c]{@{}c@{}}Only  Fg\end{tabular} 
			& {49.15}				&  49.15$^*$				& ---~
			& {73.70}				&  73.70$^*$ 			& ---~
	\\ \hline
	GoCNN
			& \textbf{50.39}			& 49.60~~				& 40.03~		
			& \textbf{75.00}			& 74.21~~				& 66.98~
	\\ \hline
	\end{tabular}
	}
	 \vspace{0mm}
\end{table*}

\FloatBarrier

\subsection{Training Models With Complete Privileged Information}
\renewcommand{\arraystretch}{1.2}
\begin{table}[t]
\setcounter{table}{0}
\centering
	\caption{Validation accuracy (for 10-crop validation) of different models on 
			 ImageNet validation set. All the compared models are trained on the ``ImageNet-0.1m'' dataset with complete privileged information. We use the ResNet-18 as the basic CNN model.}
		 \label{ImageNet-Complete}
	\vspace{-2mm}	
	{
	\small
	\begin{tabular}{c|c|c|c}
	\hline
			& \multicolumn{3}{c}{Top-1 (Top-5) Accuracy (\%)}                                                                                                                         
	\\ \hline
	        & \begin{tabular}[c]{@{}c@{}}Main\_classifier\end{tabular} 
	        & \begin{tabular}[c]{@{}c@{}}Fg\_classifier\end{tabular} 
	        & \begin{tabular}[c]{@{}c@{}}Bg\_classifier\end{tabular} 
	\\ \hline
	SVM+	
			& {37.53 (---)} 		& 	---						& ---
	\\
	ResNet-18	
			& {46.00 (70.05)}		& 	---						& ---
	\\  \hline
	GoCNN
			& \textbf{50.39 (75.00)}			& 49.60	(74.21)				& 40.03	(66.98)
	\\ \hline
	\end{tabular}
	}
	\vspace{-2mm}	
\end{table}
\setcounter{table}{2}

In this subsection, we consider the scenario where every training sample has complete privileged segmentation information. 

Firstly, we evaluate the classification performance of GoCNN on the ``ImageNet-0.1m'' dataset. Table~\ref{ImageNet-Complete} summarizes the accuracy of different models. As can be seen from the first column (\textit{Main\_classifier}), given the complete privileged information, our proposed GoCNN presents stronger generalization ability than compared models. Within the GoCNN, the \textit{Main\_classifier} has a higher validation accuracy compared with the \textit{Fg\_classifier} indicating that combining the background information can benefit object recognition to some extent. 
To further investigate the contribution of each component within GoCNN to final performance, we conduct ablation experiments and summarize the results in Table~\ref{ImageNet-Complete2}. In the experiments, we purposively prevent the gradient propagation from the other components other than the one being investigated during training. In addition, we evaluate the baseline method in a new setting where the background content is removed and only the foreground object is presented in each training image, denoted as \textit{ResNet-18-obj}. 
Comparing the results among different classifiers, one can see that learning background features can actually improve the overall performance. When  compare the \textit{Fg\_classifier}  of \textit{ResNet-18-obj} and \textit{Only Fg}, one can see the importance of the background in learning more robust and diverse foreground features. 

\begin{figure*}[t!]
	\center
	\includegraphics[width=0.851\textwidth]{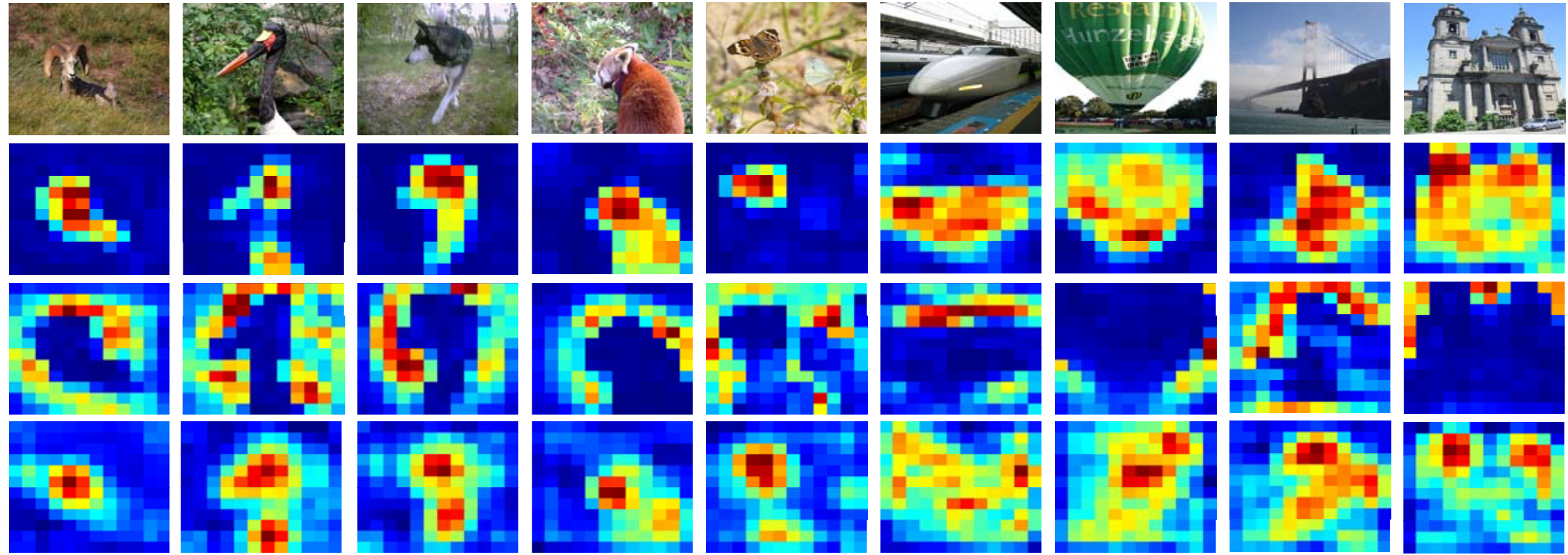}
\vspace{-1mm}	
	\caption{Visualization on the aggregated activation maps in GoCNN and vanilla ResNet-18, where hotter color indicates stronger activation. Top row: input images. Second row: representations learned by GoCNN from foreground. Third row: representations learned by GoCNN from background. Bottom row: representations learned from ResNet-18. }
	\vspace{-1mm}	
	\label{imagenet_visualize}
\end{figure*}

Secondly, to more intuitively understand  effectiveness of the proposed scheme that  learns representations  in two orthogonal groups, we visualize the activations within each group of feature maps of \emph{testing} images in Figure~\ref{imagenet_visualize}. The feature maps are generated by the final convolution layer from testing images of  $384 \times 384$ resolution. We aggregate feature maps within the same group into one feature map by taking the maximum across different feature maps (analogous to maxout operation~\cite{goodfellow2013maxout}). As it can be seen from Figure~\ref{imagenet_visualize}, the learned foreground and background representations are separated and uncorrelated within GoCNN. Compared with the baseline model, more locations get activated and richer features are detected in each group of GoCNN. The visualized results are close to  foreground/background segmentation. It is however worth emphasizing that the loss functions in GoCNN are different from the segmentation one. As shown in the second row and third row, GoCNN only penalizes the overlapped features and does not require every pixel within the target area to be activated. GoCNN is specifically designed for classification tasks in this application.

Finally, we further evaluate the generalization ability of GoCNN on the PASCAL VOC dataset. It is well known that an object may share some common properties with others even if they are  from different categories. In this experiment, we fine-tune all the evaluated models on the PASCAL VOC images to test whether the learned features from ImageNet can generalize well to a new dataset. The evaluation results are given in Table~\ref{VOC-Result}, and our proposed GoCNN shows higher average precision across all categories. It demonstrates the GoCNN learns more representative and diverse features so that they are easier to generalize from one dataset to another.

\renewcommand{\arraystretch}{1.4}
	\vspace{-1mm}	
\begin{table*}[]
	\centering
	\caption{Classification results on PASCAL VOC 2012 (train/val). The performance is measured 
			 by Average Precision in \%.
			 }	
	\vspace{-2mm}		 
	\label{VOC-Result}
	\resizebox{1\textwidth}{!}
	{
	\begin{tabular}{c|cccccccccccccccccccc|c}
		
		\hline 
		Model 
		& areo   & bike   & bird   & boat   & bottle & bus    & car    & cat    & chair  & cow
		& table  & dog    & horse  & mbk    & prsn   & plant  & sheep  & sofa   & train  & tv
		& \textbf{mean}  \\
		\hline 
		ResNet-18
		& 95.2   & 79.3   & 90.2   & 82.8   & 52.6   & 90.9   & 78.5   & 90.2   & 62.3   & 64.9
		& 64.5   & 84.2   & 81.1   & 82.0   & 91.4   & 50.0   & 78.0   & 61.1   & 92.7   & 77.5
		& 77.5  \\
		GoCNN
		& \textbf{96.1}   & \textbf{81.0}   & \textbf{90.8}   & \textbf{85.3}   & \textbf{56.0}   
		& \textbf{92.8}   & \textbf{78.9}   & \textbf{91.5}   & \textbf{63.6}   & \textbf{69.7}
		& \textbf{65.1}   & \textbf{84.8}   & \textbf{84.0}   & \textbf{83.9}   & \textbf{92.3}   
		& \textbf{52.0}   & \textbf{83.9}   & \textbf{64.2}   & \textbf{93.8}   & \textbf{78.6}
		& \textbf{79.4}   \\[0.5ex]
		\hline 
		
	\end{tabular}
    }
    \vspace{-3mm}
\end{table*}
\renewcommand{\arraystretch}{1.2}

\subsection{Training GoCNN with Partial Privileged Information}

In this subsection, we investigate the performance of different models  using partial privileged information. The experiment is conducted on the ``ImageNet-0.1m'' and ``ImageNet-Full'' datasets. We use the ``ImageNet-0.1m'' to evaluate the performance of the GoCNN by varying the percentage of privileged information (\emph{i.e.}, percentage of training images with segmentation annotations) from $20\%$ to $100\%$. The ``ImageNet-Full'' dataset is used to test the performance  of GoCNN on very large training dataset with more complex CNN architecture.

For the ``ImageNet-0.1m'' dataset, the results are shown in Table~\ref{ImageNet-Percentage}. From the results, one can observe that  increasing percentage of privileged information will  monotonically increase the performance of GoCNN until the percentage of privileged information reaches $80\%$. The performance on increasing the percentage from $40\%$ to $100\%$ is only $0.71\%$ compared with $0.92\%$ on the increasing from $20\%$ to $40\%$. This is probably because the suppression losses are more effective than we expected; that is, with very little guidance from the privileged information, the network can already be able to separate foreground and background features space and explore new features within each group.

We further evaluate  the performance of GoCNN on full ImageNet dataset with only 10\% privileged information. Here, we use the 152-layer ResNet as our basic model. As can be seen from Table~\ref{ImageNet_complete}, the GoCNN achieves 21.8\% top-1 error while the vanilla ResNet-152 has 23.0\% top-1 error. Such performance boost is consistent with the results shown in Table~\ref{ImageNet-Percentage}, which again confirms the effectiveness of the GoCNN.

\begin{table}[h]
	\centering
	\caption{Top-1 validation accuracy (single crop, in \%) with 20\%--100\% privileged information on ``ImageNet-0.1m'' dataset. Since the baseline method (ResNet-18) does not use 
			 privileged information, its validation accuracy remains the same across different tests. }
	\label{ImageNet-Percentage}
	\vspace{-2mm}	
	{
	\begin{tabular}{c|lllll}
		
	\hline 
		
	\multicolumn{1}{c|}{Model} 
	& \multicolumn{1}{c}{20\%}  		& \multicolumn{1}{c}{40\%}
	& \multicolumn{1}{c}{60\%}  		& \multicolumn{1}{c}{80\%}
	& \multicolumn{1}{c}{100\%}
	\\ 
		
	\hline 
	
	ResNet-18  	&  44.3  			&  44.3	 			&  44.3   
				&  44.3	  			&  44.3
	\\
	GoCNN	    &  \textbf{47.0}	&  \textbf{47.9}	&  \textbf{48.2}
				&  \textbf{48.6}	&  \textbf{48.6}
	\\
		
	\hline 
	
	\end{tabular}
    }
	\vspace{-3mm}	
\end{table}

\begin{table}[h]
	\centering
	\caption{Validation accuracy (for single crop validation, in \%) with 10\% privileged information on ``ImageNet-Full'' dataset.}
	\label{ImageNet_complete}
	\vspace{-2mm}	
	\resizebox{0.48\textwidth}{!}
	{
	\begin{tabular}{c|cc}
		
	\hline 
		
	\multicolumn{1}{c|}{Model} 
	& \multicolumn{1}{c}{Top-1 Accuracy}
	& \multicolumn{1}{c}{Top-5 Accuracy}
	\\ 
		
	\hline 
	ResNet-101 \cite{he2015deep} 	&  76.4  			&  92.9	
	\\	
	ResNet-152 \cite{he2015deep} 	&  77.0  			&  93.3	
	\\
	GoCNN (152 layers)  			    &  \textbf{78.2}	    &  \textbf{93.9}
	\\
		
	\hline 
	
	\end{tabular}
    }
\end{table}

\section{Discussions}
Based on our experimental results, we provide answers to the following two important questions.

\emph{Does background information indeed help object recognition for deep learning methods?}
Based on our experiments, we give a positive answer. Intuitively, background information may provide some ``hints'' for object recognition. However, though several works~\cite{song2011contextualizing,russakovsky2012object} have proven the usefulness of background information when using handcraft features, few works have studied the effectiveness of background information on deep learning methods for object recognition tasks. Based on the experimental results shown in Table~\ref{ImageNet-Complete2}, both the foreground classification accuracy and the overall classification accuracy can be further boosted with our proposed framework. This means that the background deep features can also provide useful information for foreground object recognition.

\emph{Can a more precise annotation with richer information, \emph{e.g.}, segmentation annotation, assist the image classification training process?} 
The answer is clearly yes. In fact, in recent years, several works have explored how object detection and segmentation can benefit each other~\cite{dai2015boxsup,hariharan2014simultaneous}. However, none of existing works has studied how image segmentation information can help train a better classification deep neural network. In this work, by treating the segmentation annotations as the privileged information, we first demonstrate a possible way to utilize segmentation annotations to assist image classification training.

\vspace{-1mm}	
\section{Conclusion}
We proposed a group orthogonal neural network for image classification which 
encourages learning more diverse feature representations. Privileged information is 
utilized to train the proposed GoCNN model. To the best of our knowledge, we are the first to 
explore how to use image segmentation  as  privileged information to assist CNN training for image classification. 

\vspace{-1mm}	
\section*{Acknowledgments}
The work of Jiashi Feng was partially supported by National University of Singapore startup grant R-263-000-C08-133 and Ministry of Education of Singapore AcRF Tier One grant R-263-000-C21-112.

\appendix

\bibliographystyle{named}
\bibliography{ijcai17}

\end{document}